\documentclass{article}
\usepackage{ijcai22}
\usepackage{bbm}
\usepackage{times}
\usepackage{soul}
\usepackage{url}
\usepackage{amssymb}
\usepackage[hidelinks]{hyperref}
\usepackage[utf8]{inputenc}
\usepackage[small]{caption}
\usepackage{graphicx}
\usepackage{amsmath}
\usepackage{amsthm}
\usepackage{booktabs}
\usepackage{algorithm}
\usepackage{algorithmic}
\newtheorem{problem}{Problem}
\title{Dynamic Graph Learning Based on Hierarchical Memory for \\ Origin-Destination Demand Prediction}
\author{
Ruixing Zhang\and
Liangzhe Han\and
Boyi Liu\and
Jiayuan Zeng\And
Leilei Sun\footnote{Contact Author}\\
\affiliations
SKLSDE Lab, Beihang University, Beijing 100191, China\\
\emails
\{yyxzhj, liangzhehan, 1906boy, yuaner, leileisun\}@buaa.edu.cn}
\begin{document}
\maketitle
\begin{abstract}
Recent years have witnessed a rapid growth of applying deep spatiotemporal methods in traffic forecasting. However, the prediction of origin-destination (OD) demands is still a challenging problem since the number of OD pairs is usually quadratic to the number of stations. In this case, most of the existing spatiotemporal methods fail to handle spatial relations on such a large scale. To address this problem, this paper provides a dynamic graph representation learning framework for OD demands prediction. In particular, a hierarchical memory updater is first proposed to maintain a time-aware representation for each node, and the representations are updated according to the most recently observed OD trips in continuous-time and multiple discrete-time ways. Second, a spatiotemporal propagation mechanism is provided to aggregate representations of neighbor nodes along a random spatiotemporal route which treats origin and destination as two different semantic entities. Last, an objective function is designed to derive the future OD demands according to the most recent node representations, and also to tackle the data sparsity problem in OD prediction. Extensive experiments have been conducted on two real-world datasets, and the experimental results demonstrate the superiority of the proposed method. The code and data are available at https://github.com/Rising0321/HMOD. 

\end{abstract}

\section{Introduction}

In the past decade, with the rapid growth of traffic infrastructure, an extensive amount of high-quality traffic data is collected for the industry and the research community.
This opportunity promotes multiple promising applications of Intelligent Transportation System (ITS) including traffic time estimation~\cite{han2021multi}, traffic trajectory prediction~\cite{ma2019trafficpredict}, and traffic demand prediction~\cite{yao2018deep}.

Almost all the traffic applications require a meaningful and successful representation for traffic nodes (e.g., road segments, metro stations), which involves spatial dependency among nodes and temporal dependency across time.
Among these applications, traffic demand prediction, which has a significant influence on transportation planning, is a vital and attractive domain in ITS.
Especially, the node-to-node OD demand prediction attracts more and more researchers in recent years since it can not only provide the number of passengers, but it can also show which target places passengers tend to choose, which implies a potential to better model human mobility and further benefit the whole system.

To solve the problems, researchers have proposed a variety of models based on cutting-edge deep learning methods, such as CNNs~\cite{noursalehi2021dynamic}, RNNs~\cite{zhang2021short} and Transformers~\cite{sankar2020dysat}.
GNNs, as their potential to learn high-quality representation on graph, are utilized in OD demand prediction to mine the complex spatial dependencies, which has achieved considerable success~\cite{wang2019origin};~\cite{shi2020predicting};~\cite{wang2021gallat}.

Despite all the efforts, two important issues are rarely discussed:
First, previous methods tend to divide the dynamic trips into multiple regular-spaced time slices, using OD matrix at each time slice to represent the demand during the corresponding time(e.g. \cite{zhang2021short}).
This process makes it easy to fit existing recurrent deep learning methods, but this data compression replaces raw time information with discrete-time slices and misses useful information.
Second, previous methods consider node relations directly and simply.
For example, some researchers view two nodes are closely related if there are a great number of passengers from one to another in the last period like \cite{shi2020predicting}.
However, they fail to distinguish between origins and destinations, which neglects relations between origin nodes that have similar demands. 

This paper aims to address the above two issues and proposes a dynamic node representation framework for OD demand prediction.
However, it is a nontrivial endeavor to design such a framework due to the following challenges: 
First, the time granularity about how to organize the historical OD demand matrix is hard to choose. 
Choosing too coarse time granularity will result in an inability to sense useful information such as the trend while choosing too fine time granularity will lead to substantial noise.
How to handle the complex temporal dynamics of trips is fundamental to node representations.
Second, relations between nodes are time-evolving.
Demand from two nodes to other nodes could be similar on weekdays and completely different on weekends.
How to dynamically capture these relations is crucial to node representations.
Third, the model is required to predict demand values on every pair of nodes.
Due to demand imbalance, many OD pairs have no demand at a certain time.
How to properly handle these pairs is also challenging to stable training.

In this paper, we propose a Hierarchical Memory dynamic graph representation learning framework for OD Prediction (HMOD).
First, a hierarchical memory updater is designed to integrate the continuous-time information and multi-level discrete-time information.
This updater avoids limits of time information loss in previous discrete-time methods.
Meanwhile, it constantly updates memory for dynamic node representation by message passing scheme and a message fusion mechanism is provided to integrate hierarchical memories.
Second, a spatiotemporal propagation module is designed to aggregate neighbor node representations.
This module is based on real-time demand and treats origins and destinations as two different semantic entities.
It can effectively exploit similar relations between origins instead of a mix of origins and destinations.
Third, an output layer and a specially designed objective function are deployed to predict the OD matrix.
The objective function relaxes the error of no-demand pairs to alleviate the influence of data imbalance.


The contributions of this work are summarized as follows:
\begin{itemize}
\item We design hierarchical memories to integrate discrete-time information and continuous-time information of OD demand. To the best of our knowledge, it is the first time that traffic node representation learning is extended to the continuous-time dynamic graph view.

\item We propose an origin-destination embedding module to aggregate neighbor information along conditioned random walks. The module views origins and destinations as different semantic entities and can thus distinguish them to avoid mixing two types of information up.

\item Extensive experiments are conducted on two real-world traffic datasets to evaluate the performance of the proposed framework and key components. The results demonstrate that our framework significantly outperforms other state-of-the-art methods.
\end{itemize}



\section{Related Work}

\subsection{OD Demand Prediction}

With extensive trip data are recorded by traffic cards and APPs
, researchers are attracted to solve a more fine-grained traffic demand prediction, OD demand prediction. A simple solution is directly taking OD matrix as a two-dimensional image and applying convolution-based or spectral-based methods on it, such as  ~\cite{noursalehi2021dynamic}. This kind of method fails to capture the spatiotemporal dependencies because they are limited to consider further relationship and topology, which is complex but common in traffic. Along with the development of graph neural networks, some researchers view stations or zones as nodes in the graph and design a variety of frameworks to address the problem of learning spatiotemporal dependencies. GEML~\cite{wang2019origin} uses GNN and LSTM to perform grid representation and capture temporal patterns. DNEAT~\cite{zhang2021short} design a spatiotemporal attention network and exploit different time granularity to mine complex temporal patterns. Although amazing results have been achieved by these methods, they still face the problem of missing fine-grained patterns since they are all based on discrete-time snapshots to update the representation.

\subsection{Dynamic Graph Representation Learning}

Dynamic graph representation learning is a technique to maintain evolving node states with time going on, while traditional graph representation learning frameworks generate static representations ignoring the dynamic changing content. For example, GraphSAGE~\cite{hamilton2017inductive} and GAT~\cite{velivckovic2017graph} train an unchanged embedding without considering inserted or removed edges.

Recently, Dynamic Graph Representation Learning methods are proposed to address the above issues. Generally, the methods can be divided into two categories. The first category is called Discrete-Time Dynamic Graph (DTDG). The DTDG methods firstly define a length $\tau$ and then will update the embeddings every $\tau$ unit time. DynamicTraid~\cite{zhou2018dynamic} and  tNodeEmbed~\cite{singer2019node} are the representative ones. Although previously mentioned OD prediction methods are all able to model evolving patterns of node representations, it faces severe problems to sense fine-grained information. On the contrary, the second category, named Continuous-Time Dynamic Graph (CTDG), updates node representations when an event happens. It is a more natural way to update the embeddings since raw events come in sequence instead of coming as snapshots. The typical ones are TGN~\cite{rossi2020temporal}, DyRep~\cite{trivedi2019dyrep}, JODIE~\cite{kumar2019predicting}. However, these methods could only capture temporal dependencies of limited time steps.

\section{Preliminaries}

The trip dynamic graph is formulated as $\mathcal{G}$ = $(V,\ E)$, where $V = \{v_1,v_2,\cdots, v_N\}$ is the set of $N$ nodes and $E$ = $\{e_1,e_2,\dots,e_M\}$ is the edge set of $M$ edges.
These nodes represent stations or regions in the real world. 
Each edge $e_m = (u_m,v_m,t_m, \bf{f_m})$ represents a passenger departed at $t_m$ from $u_m$ to $v_m$ with associated feature vector ${\bf{f_m}} \in \mathbb{R}^{d_F}$. 
The feature vector contains travel distance, weather, payment, and tips. 
At a specific time point $t$, the dynamic graph can be represent as $\mathcal{G}_t = (V, \{e_m|t_m<t\})$, which contains trips happened before $t$. 
The OD demand matrix during time interval from $t$ to $t+\tau$ is denoted is $\mathbf{Y}_{t:t+{\tau}} \in \mathbb{R}^{N*N}$ whose (i,j)-entry represents the volume from $v_i$ to $v_j$ during this time interval. 
The purpose of OD demand prediction is to predict future OD matrix $\mathbf{Y}_{t,t+{\tau}}$ given historical trip data as a dynamic graph $\mathcal{G}_t$.

\begin{problem}
Given the dynamic graph $\mathcal{G}_t$ at timestamp $t$, it is expected to learn a function $\Psi$ to predict the OD Matrix 
$\hat{\mathbf{Y}}_{t,t+{\tau}}$  = $\Psi(\mathcal{G}_t) $, i.e.,%
\begin{align}
   \Psi^* =\mathop{\arg\min}\limits_{\Psi} {\rm ODLoss}( \mathbf{Y}_{t,t+{\tau}},\Psi(\mathcal{G}_t) ),
\end{align}%
where ODLoss will be given in the next section.
\end{problem}

\begin{figure*}[t]
\centering
\includegraphics[width=1.0\textwidth]{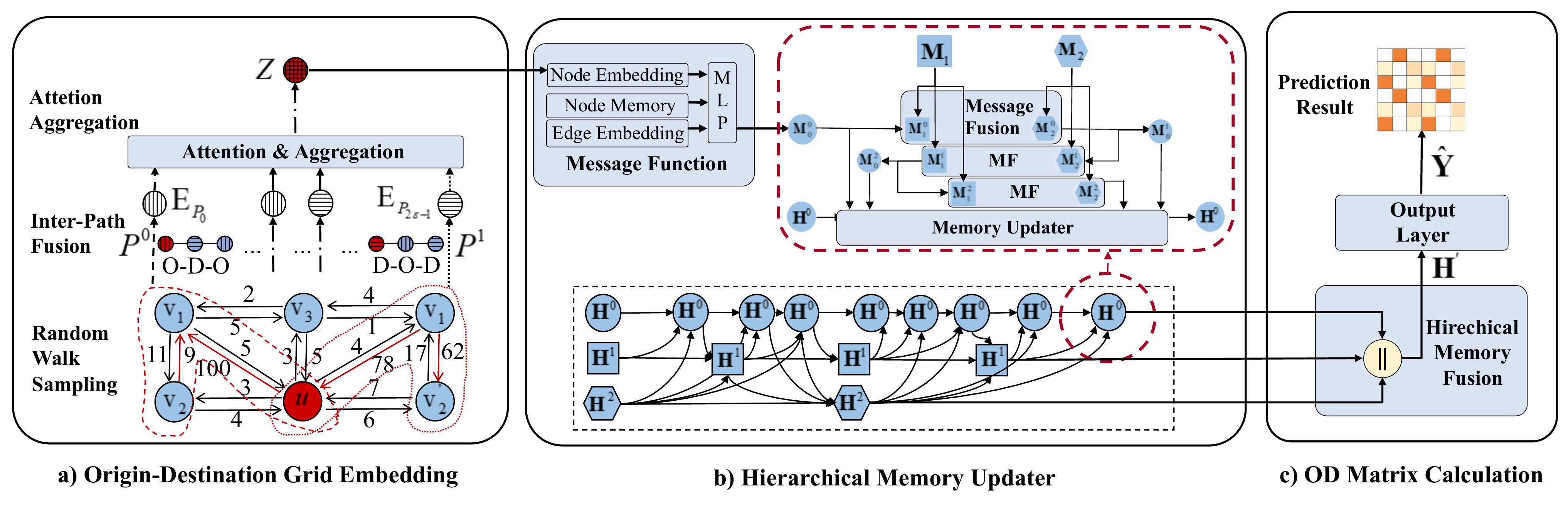} 
\caption{The Overall Framework of HMOD.}
\label{fig1}
\end{figure*}
 
\section{Methodology}
In this section, we will elaborate details of {\bf HMOD}.
The overall architecture is shown in Figure \ref{fig1}.
First, we will introduce the hierarchical memory updater which maintains the node representations and dynamically update them based on message passing, which can preserve hierarchical time semantic information for nodes. 
Next, to generate high-quality messages for memory update, an Origin-Destination Node Embedding method is designed to aggregate neighbor information by spatiotemporal random walks. 
In the last part, the output layer and specially designed objective function are introduced for the final prediction and training.

\subsection{Hierarchical Memory Updater}
Historical trips, which are original edges with continuous timestamps, reflect OD demand and are fundamental to node representation.
Previous methods like GEML generally count edges in a fixed time window and make its time discrete, which discards practical time information.
The more coarse time granularity is to generate time windows, the more information is lost.
Unlike previous methods, we propose to maintain hierarchical-time memories for nodes by a hierarchical memory updater to address this problem.
On the one hand, a part of it views time as continuous features and is updated when an edge comes.
On the other hand, another part maintains multiple discrete-time memories to reveal a macro node state.
Further, a memory fusion mechanism is designed to mix each level's memory while keeping mainly receiving the corresponding level message. Formally, $D + 1$ memories 
${\bf H }\in {\mathbb{R}^{(D+1) \times d_H}}=\{{\rm \bf H}^0, {\rm \bf H}^1, {\rm \bf H}^2, \cdots, {\rm \bf H}^{\rm D}\}$ 
are maintained on each node, where ${\rm \bf H}^0$ represents continuous-time memory
and ${\rm \bf H}^{\rm d} (d\geq1)$ represents macro discrete-time memory covering 
$\Delta \rm T_d$ time unit. Their updating methods are as follows.


\subsubsection{Message Function}
The above hierarchical memory aims to always maintain updated node states.
When a memory updating process is triggered, that is to say for continuous-time memories there is a new edge coming and for discrete-time memories, $\Delta \rm T_d$ time unit passed by, we need to calculate a message to update node memories according to information triggered this update.
For continuous-time memories, the message contains the last updated node states, encoding of the time gap from the last update, and aggregated neighbor information which will be discussed in 4.2. 
For discrete-time memories, the message contains the last updated node states, OD matrix during $\Delta \rm T_d$, and aggregated neighbor information.
For the specefic node i,
the above message generation process can be formulated as:
\begin{align}
&{\rm \bf M' } =\left\{
\begin{aligned}
\ [{\rm \bf H}^d||{\rm\Theta  ({\rm \bf H}^d,{\mathcal{G}_t})}||{\rm \Omega_d({\rm \bf H}^d,{\mathcal{G}_t})}],d=0\\
\ [{\rm \bf H}^d||{\rm \mathbf{Y}_{t-2^{\rm  d-1} \Delta \rm T,t}^i}||{\rm \Omega_d({\rm \bf H}^d,{\mathcal{G}_t})}],d\neq0\\
\end{aligned},
\right.\\
&{\rm \bf M}^0_d = {{\bf W}^{m_1}\sigma({\bf W}^{m_2}{\bf M}'+{\bf b }^{m_2})+{\bf b }^{m_1}},\\
&{\rm \Theta}({\rm \bf H}^d,\mathcal{G}_t) = (1+{\bf W}^e(t-{\bf t}_i^-)+{\bf b}^e)\odot{\rm \bf H}^d,
\end{align}%
where ${\rm \bf M}^0_d \in {\mathbb{R}^{d_M}}$ represents the message described above, $\odot$ represents element-wise product, ${\bf t}_i^-$ represents last update time of node i, ${\rm \mathbf{Y}_{t-2^{\rm  d-1} \Delta \rm T,t}^i}$ means the i-th row of the OD matrix.  ${\bf W}^*$ and ${\bf b}^*$ are learnable parameters. ${\rm \Theta}({\rm \bf H}^d,{\rm \bf G})$ represents the time encoding, which is utilized to sensing different time span's different effect. $\Omega_d({\rm \bf H}^d,{\rm \bf G})$ represents the node embedding method. $\sigma$ represents the ReLU function. 

\subsubsection{Message Fusion}

The message directly computed by the previously mentioned message function only contains information on a specific track, i.e. continuous information or discrete information of a specific granularity. 
However, the continuous message may lack certain global spatiotemporal information while the fine-grained spatiotemporal information may be absent in discrete messages. However, how to integrate other time granularity information while preserving own message is a challenging problem. To this end, a message fusion mechanism is proposed to integrate the specific track message with others to integrate multiple types of spatiotemporal information.

Formally, the latest message ${\bf M} \in {\mathbb{R}^{(L+1) \times d_M}} = \{{\bf M}^0,{\bf M}^1,\cdots,{\bf M}^L\}$ is stored with corresponding memory, where ${\bf M}^0$ is the output of message function and ${\bf M}^l (l\geq 0)$ is computed as follows. When computing  ${\bf M}^l$, other messages from other time level is first aggregated then concatenated with ${\bf M}^{l-1}$. It is intended to mix information from other time granularity while maintain the corresponding  time granularity information. The concatenation is then passed to a fully connected layer to calculate the result. Intuitively, if treating the updating process of a node's memories as a graph where each memory at a certain timestamp is a node, ${\bf M}^0$ represents node's feature and ${\bf M}^l$ aggregate ${\bf M}^{l-1}$ with its l-hop neighbor. This mechanism can be formulated as:

\begin{align}
{\rm \bf M}_d^l ={\bf W}_{mess}[{\rm \bf M}_d^{l-1} || {\rm AGG}(\{{\rm \bf M}_k^{l-1},\forall  k\neq d \})],
\\
{\rm AGG}(\{{\rm \bf M}_t\})  = 
{\rm max} (\{{\sigma(\bf W}^{pool}{\rm \bf M}_t+{\bf b}^{pool})\}),
\end{align}%
where ${\rm \bf M}_d^l$ refers to message of $d^{th}$ memory at $l^{th}$ layer, $\sigma$ is ReLU function, ${\bf W}^{pool}$, ${\bf W}^{mess}$, ${\bf b}^{pool}$ and ${\bf b}^{mess}$ are learnable parameters. Besides, a visual explanation of the implication graph is present in Figure\ref{fig1} b).

\subsubsection{Memory Updater}

When the model has computed the message for each memory, another problem arises: how to effectively update memories while preventing the model from forgetting past information. Thus, due to the capability to memorize information with reduced parameters, GRU~\cite{cho2014learning} is employed to update memories here. On account of a deeper message having more information, ${\rm \bf M}_d^L$ is directly used as GRU's input, which can be formulated as:
\begin{align}
{\rm \bf H }^d = {\rm GRU}({\rm \bf H }^d,{\rm \bf M}^L_d).
\end{align}%

Note that there exist extremely dense edges in OD prediction, for example, millions of edges may occur in a day.
It would take too much time for the model to train with one edge each iteration. 
Thus, a message aggregator is designed to aggregate continuous-time messages of an edge batch and reduce the training time. 
To be specific, instead of simply average messages of multiple edges, we extend the last update memory with another item which both considers node states and influence of time in a batch as exponential decay.
Then we replace the first $\bf H$ in ${\bf M}'$ with the average of the concatenation memories in a batch. Formally, when $d=0$, for the specific node i, equation (2) can be rewritten as:
\begin{align}
&{\rm \bf H''} = {{\bf MEAN}_{b \in B }([{\rm \bf H}^0||{\rm \bf H}^0*{\bf exp}(t-{\bf t}_b^-)])},\\
&{\rm \bf M' } =\left\{
\begin{aligned}
\ [{\rm \bf H}''||{\rm\Theta  ({\rm \bf H}^d,{\rm \bf G})}||{\rm \Omega_d({\rm \bf H}^d,{\rm \bf G})}],d=0,\\
\ [{\rm \bf H}^d||{\rm \mathbf{Y}_{t-2^{\rm  d-1} \Delta \rm T,t}^i}||{\rm \Omega_d({\rm \bf H}^d,{\rm \bf G})}],d\neq0,\\
\end{aligned}
\right.
\end{align}%
where $b$ represents edges departed at i in a batch. Note that the choice of the batch can be variant. The message of continuous memories can be aggregated every $\rho $ edges while being aggregated every $\tau$ time unit is also acceptable. The latter is chosen in our implementation for convenience.

\subsection{Origin-Destination Node Embedding}
When embedding traffic nodes to feature space, it is crucial to consider spatial topology relations between nodes.
Most previous works for OD demand prediction capture nodes' spatial relations in a direct and simple way; they mostly leverage simple GNNs on either predefined geographic graph or last transition graph.
In summary, they treat origin nodes and destination nodes as the same semantic entities.
However, it may not fit this problem to neglect different semantic meanings between the origin nodes and destination nodes.
Besides, since the OD graph is a fully connected graph, simply aggregating all neighbors of a node may result in high computation complexity. 
Therefore, a random walk based Origin-Destination Node Embedding is proposed for an effective and efficient node embedding. 

\subsubsection{Random Walk Sampling}
Random walk-based methods have achieved amazing success in graph embedding due to their ability to sense graph structure, e.g. DeepWalk~\cite{perozzi2014deepwalk}, node2vec~\cite{grover2016node2vec}. 
However, directly utilizing previous random walk methods will mix up semantic relations of origin nodes and destination nodes, which should be carefully considered in OD demand prediction. 
For the example in Figure \ref{fig1} a), there exists a dense OD demand from $v_3$ to $v_4$ and from $v_3$ to $u$ at morning peak, where $u$ and $v_4$ are two working area and $v_3$ is a living area.
It appears that though there is little demand between $u$ and $v_4$, they are more likely to share a similar demand pattern to other nodes. 
Motivated by this, the following random walk procedure is designed to distinguish the semantic entities while aggregating neighbor information.


This procedure can be explained as a) keeping sampling forward edges and reverse edges alternately and recursively, b) keeping sampling reverse edges and forward edges alternately and recursively. 
For example, when calculating embedding for node $u$ in Figure \ref{fig1} a), a random sampling process first starts from OD forward edges with origin node $u$ and get destination node $v_1$. Then, OD reverse edges with destination $v_1$ are random sampled to get origin node $v_2$. Finally, a node-set with $\omega$ nodes is sampled, which is denoted as $P^0$.
For another random walk, a random sampling process starts from OD reverse edges with destination node $u$ and gets origin node $v_3$. 
Then, edges with origin node $v_3$ are random sampled to get node $v_4$. 
Finally, a node-set with $\omega$ nodes is sampled, which is denoted as $P^1$. 
Repeatedly using the above two random walk strategies, $2\varepsilon$ walks are obtained. 
For each step, the random sample probability is formulated as:
\begin{align}
P = \left\{
\begin{aligned}
\phi(\exp(\hat t-{\bf t}^-)) ,d=0,\\
\phi(\rm Y_{t-2^{\rm  d-1} \Delta \rm T,t} + \epsilon),d\neq 0,\\
\end{aligned}
\right.
\end{align}%
where $\phi$ represents row-wise normalization to make the sum of probability equal to 1, $d=0$ represents generating random walks according to continuous-time edges and $d\neq 0$ represents generating random walks according to $d^{th}$ discrete-time memories.
When calculating messages for continuous-time memories, $\hat t$ is firstly initialized as the calculating time $t$. When sampled an edge $(u,v,t')$,  $\hat t$ is changed to t'.

\begin{table*}
\setlength\tabcolsep{3pt}
\renewcommand{\arraystretch}{1.1}
\centering
\begin{tabular}{|c|cc|cc|cc|cc|cc|cc|}
\hline
& \multicolumn{6}{|c|}{Beijing Metro} & \multicolumn{6}{|c|}{New York Taxi}\\ 

\hline
  & \multicolumn{2}{|c|}{$\geq 0$} & \multicolumn{2}{|c|}{$\geq 3$} & \multicolumn{2}{|c|}{$\geq 5$}& \multicolumn{2}{|c|}{$\geq 0$}& \multicolumn{2}{|c|}{$\geq 3$}& \multicolumn{2}{|c|}{$\geq 5$}\\
\hline
Method  & RMSE & PCC& RMSE & PCC& RMSE & PCC& RMSE & PCC & RMSE & PCC& RMSE & PCC\\
\hline
HA       &4.8672  & 0.8035   & 10.5651  & 0.8012  & 13.2245  & 0.7956  & 1.4501 & 0.8481   & 3.8801  & 0.7460  & 4.8517  & 0.7140\\
LR       &5.3556  & 0.7521    & 11.7136  & 0.7325  & 15.1065 & 0.7229   & 1.3631     &  0.8586  &3.3003  & 0.8042  & 4.1724  & 0.7887\\
XGBoost    & 5.7716  & 0.7039  & 12.9623  &  0.6453  & 16.8519  & 0.6142  &  1.3575  & 0.8599  &  3.3001  &0.8064  & 4.1736  & 0.7915\\
GEML   & 4.6143  & 0.8268   & 10.1698  & 0.8075  & 13.1405  & 0.7995  & 1.3188  & 0.8718  & 3.1293 & 0.8161  & 3.9172  & 0.7980\\
DNEAT   & 5.6378  & 0.7334   & 12.8147  & 0.6622 & 16.7077  & 0.6180  & 1.5249  & 0.8242     & 3.9687  & 0.7161  & 5.1619  & 3.6520\\
TGN   & 6.1313  & 0.6453   & 13.5933  & 0.6091  & 17.6037  & 0.6003 & 1.2947  & 0.8747  & 3.0971  & 0.8188  & 3.9104  & 0.8037  \\
\hline
HMOD   &$\bf 3.7686$  & $\bf 0.8861$   & $\bf 8.2641$  & $\bf 0.8730$  & $\bf 10.6426$  & $\bf 0.8675$  & $\bf 1.1926$  & $\bf 0.8936$    & $\bf 2.8590$ & $\bf 0.8507$ & $\bf 3.5760$  & $\bf 0.8392$\\
\hline
\end{tabular}
\caption{The Comparison Results on Beijing Metro and New York Taxi.}
\label{tab1}
\end{table*}

\subsubsection{Intra-Walk Aggregation}
Based on sampled walks from the dynamic graph, we propose an intra-path aggregation module to obtain walk representations.
As described above, the walks contain nodes with different semantic meanings. 
Therefore, a special transformation layer is designed to project nodes' into the same embedding space, which passes the node sampled as the origin to a fully connected layer $W^O$ and passes the node sampled as the destination to a fully connected layer $W^D$. After node content transformation, mean aggregator is applied to compute representation of a walk, which can be formulated as:%
\begin{align}
{\bf E}_{P_i} =\left\{
\begin{aligned}
(\sum_{j=2k} {\bf W}^O{\rm \bf H}_{P^{2t'}_j}^d + \sum_{j=2k+1} {\bf W}^D{\rm \bf H}_{P^{2t'}_j}^d)/2\omega\\
(\sum_{j=2k} {\bf W}^D{\rm \bf H}_{P^{ 2t'+1}_j}^d + \sum_{j=2k+1} {\bf W}^O{\rm \bf H}_{P^{ 2t'+1}_j}^d)/2\omega\\
\end{aligned}
\right.
\end{align}
where ${\rm \bf H}_{P^{i}_j}^d$ represents the $d^{th}$ memory of node $P^{i}_j$, which may be different from our previous definition.  $j=2k$ means even position of a walk since even position and odd position has different semantic meaning, $P^{2t'}$ means even number of the walk since even number of walks start sampling from forwarding edges while an odd number of walks start sampling from reverse edges.

\subsubsection{Inter-Walk Aggregation}
After intra-walk aggregation for each walk embedding, an attention aggregation is designed to aggregate information from different walks, which can be formulated as:
\begin{align}
&{\bf E}_{P_i}’ = {\bf W}^a{\bf E}_{P_i},\\
&\alpha_i = {\bf a } {\bf E}_{P_i}’,\\
&\beta_i = \frac {\exp( \alpha_i)} {\sum_{i=0}^{2\varepsilon-1} \exp( \alpha_i)},\\
&\bf Z= \beta_i{\bf E}_{P_i},
\end{align}
where ${\bf W}^a,\ {\bf a }$ are learnable parameters. 

\subsection{OD Matrix Prediction}

\subsubsection{Fusion and Prediction}

If requiring predicting OD Matrix at timestamp $t$, the hierarchical memories at $t$ are fused to make the prediction. Since each memory contain specific granularity information, the memories are concatenated to keep the hierarchical information. Then a MLP is adopted to make the prediction. The $i^{th}$ row of prediction result can be formulated as:
\begin{align}
&{\bf H}' = \| _{d=0}^D {\bf H}^d\\
&{\bf \hat Y}^i = {{\bf W}^{o_1}\sigma({\bf W}^{o_2}{\bf H}'+{\bf b }^{o_2})+{\bf b }^{o_1}}
\end{align}%
where ${\bf W}^{o_1}$, ${\bf W}^{o_2}$, ${\bf b}^{o_1}$, ${\bf b}^{o_2}$ are learnable parameters. 

\subsubsection{OD Loss}

One last problem is there exists a large number of zeros in OD matrix while predicting them as negative numbers are also acceptable in the real-world scene. Besides, since a heavy negative number and a slice negative number may contain different meanings, directly training them to be zero could lead to poor performance. Therefore, an OD Loss is designed based on MSE, which can be formulated as:
\begin{align}
&\mathcal{L}  = \frac 1 {|{\bf Y}|} \sum_{y\in {\bf Y}} (I(y,\hat y)(y-\hat y)^2),\\
&{I(y,\hat y)} =\left\{
\begin{aligned}
&0,y=0  \&  \hat y\leq0\\
&1,{\rm else}\\
\end{aligned}.
\right.
\end{align}

\section{Experiments}
\subsection{Dataset}
To verify the performance of our framework, extensive experiments are conducted on two real-world datasets.

{\bf Beijing Metro} Beijing Metro dataset contains railway trip data in Beijing from June to July in 2018. The first 6 weeks are selected as the training set, the next 1 week as the validation set, and another next 1 week as the test set. This dataset covers 268 stations of Beijing Metro and has over 200 million timestamped edges in total. 
 
{\bf New York Taxi} New York Taxi dataset contains taxi trip data in New York from January to June in 2019. The first 139 days are selected as the training set, the next 21 days as the validation set, and the next 21 days as the test set. This dataset covers 63 regions of New York and has over 38 million timestamped edges in total.

\subsection{Experimental Settings}
The task is to predict OD demand matrix in the next $\tau = 30$ minutes with historical trip data. Adam~\cite{kingma2014adam} is used as the optimizer with learning rate 1e-4 for Beijing Metro and 1e-5 for New York Taxi, and the learning rate is chosen from [1e-2, 1e-3, 1e-4, 1e-5, 1e-6]. 
Memory dimension $d_H$ and message dimension $d_M$ are set to 128, which is chosen from [64, 128, 256, 512].
Each experiment contains 500 epochs and an early stop strategy with 20-epoch patience is used to avoid overfitting.
The model is implemented using the PyTorch framework on a machine with Intel(R) Xeon(R) Gold 6130 CPU @ 2.10GHz and 4 Tesla T4 GPUs.
PCC (Pearson Correlation Coefficient) and RMSE (Root Mean Squared Error) are adopted as metrics to evaluate the performance of our model. 

\subsection{Baseline Methods}
\begin{itemize}
\item $HA$ (Historical Average) computes the historical average of OD demand matrix as the prediction.
\item $LR$ (Linear Regression) is a regression model which exploits linear correlations between input and output.
The input of LR contains the historical OD demand of one OD pair from the last 4 consecutive snapshots. 

\item $XGBoost$~\cite{xgboost} adopts gradient boosting tree to learn from the previous pattern. 
The input of XGBoost is organized as the same as LR.

\item $GEML$~\cite{wang2019origin} is an OD demand prediction model based on snapshots and pre-defined neighborhoods. 
It adopts graph convolution along with a skip-RNN to extract spatial and temporal patterns.

\item $DNEAT$~\cite{zhang2021short} is another OD demand prediction model based on snapshots and node-edge attention. 
It adopts graph attention along with a dilated convolution to capture spatiotemporal dependencies among discrete-time snapshots of a dynamic graph. 

\item $TGN$~\cite{rossi2020temporal} is a continuous-time dynamic graph representation learning framework. 
This framework is deployed to OD prediction task by this paper. 

\end{itemize}

\subsection{Experimental Results}

\begin{figure}[t]
\centering
\includegraphics[width=0.48\textwidth]{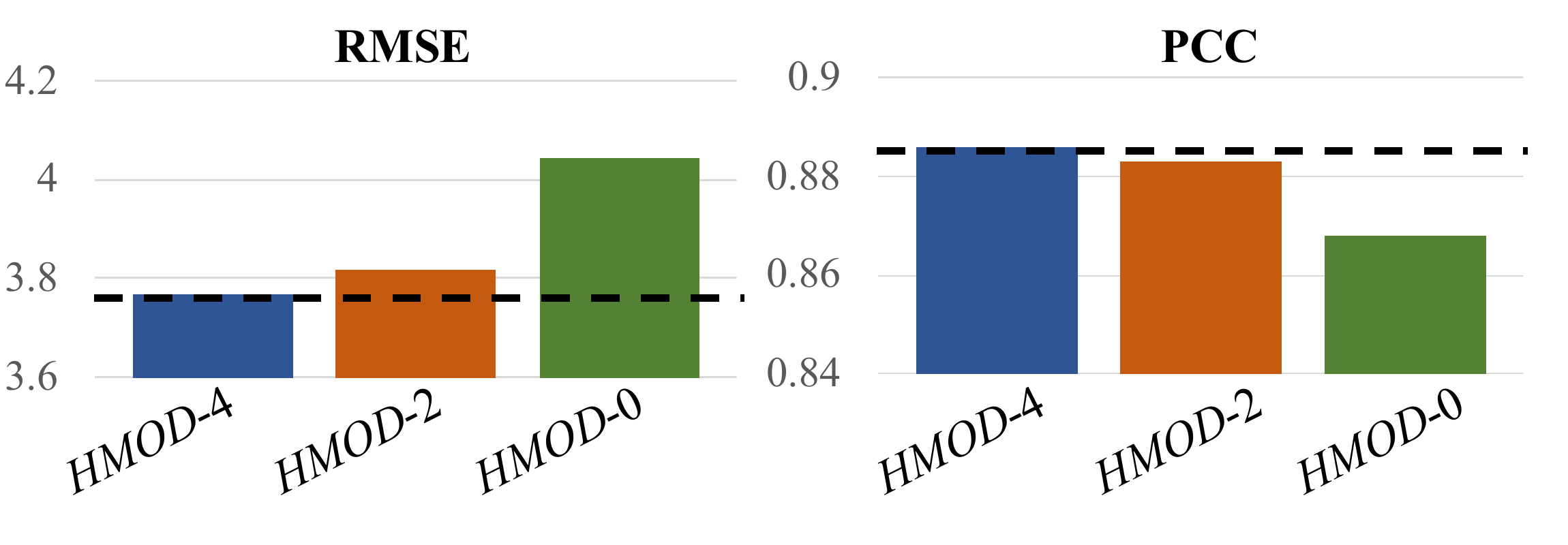} 
\caption{Ablation on Hierarchical Memories}
\label{fig2}
\end{figure}

\begin{figure}[t]
\centering
\includegraphics[width=0.48\textwidth]{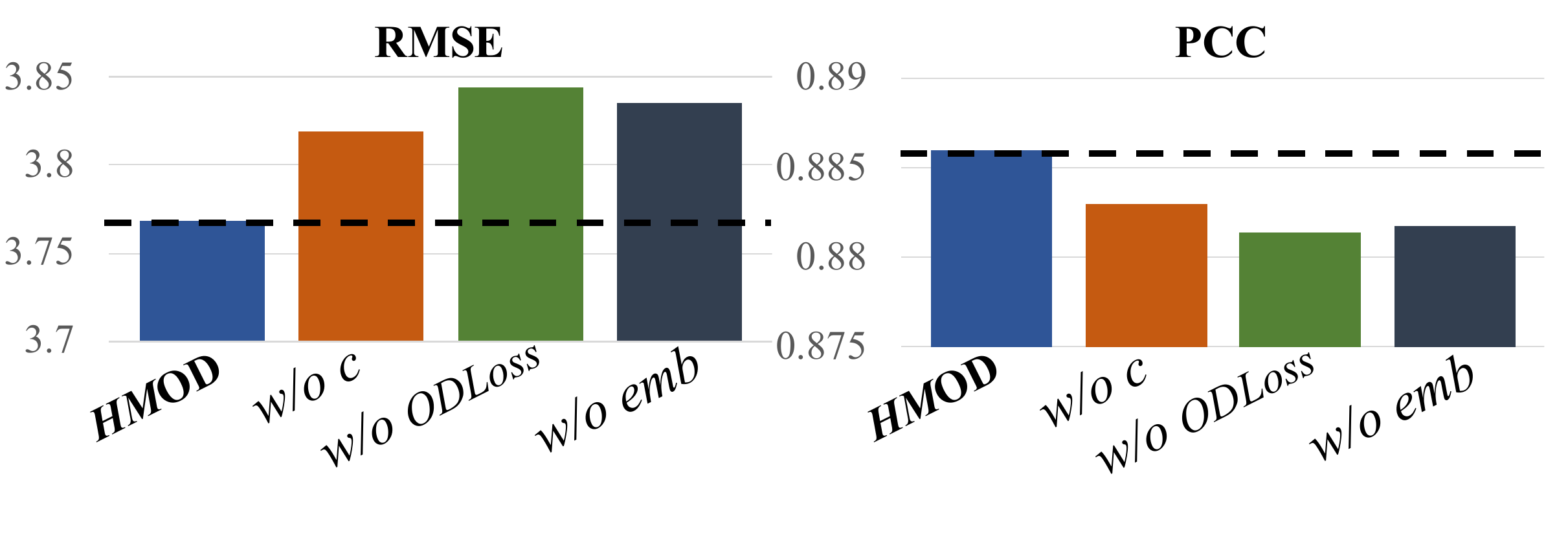} 
\caption{Ablation on Main Modules}
\label{fig3}
\end{figure}

Table \ref{tab1} shows the comparison results, where $\geq p$ means the accuracy in the case where truth-value $\geq p$.
The evaluation with different $p$ indicates whether the methods perform well in different situations.
The results show that our model outperforms all the baselines on both RMSE and PCC, both datasets, and all the situations.
In BJMetro, TGN performs poorer than other OD prediction methods.
This is because BJMetro has a larger amount nodes and edges, which may make it harder to find important neighbors for TGN.
However, in New York Taxi, it performs better than other baselines, which demonstrates the effectiveness of continuous-time information in this scenario.
Owing to hierarchical-time node representations, the proposed method could capture fine-grained and macro information simultaneously and always perform well.


\subsection{Ablation Study}
\subsubsection{Ablation on Hierarchical Memories}
The effect of hierarchical memories on Beijing Metro is shown in figure \ref{fig2}. The $\it HMOD$-x refers to our model with x discrete-time memories, while $HMOD-4$ is the setting in our model. For the convenience in our implements, the first discrete memory leverages $\delta T$ = 30 mins input, and the k-th discrete memory leverages $2^k \delta T$ mins input. The results show that only one continuous memory can achieve a significant result, which demonstrates the effectiveness of the continuous-time method. Besides, the more discrete-time information added,  the better performance will HMOD achieve, which also reflects the importance of the discrete-time information. 

\subsubsection{Ablation on Main Modules}
The impact of main modules on Beijing Metro is shown in figure \ref{fig3}. The $\it w/o\ c$ refers to removing the continuous-time memory part and containing four discrete-time memories to predict OD Matrix. The $\it w/o\ emb$ refers to removing the OD node embedding part. The $\it w/o\ ODLoss$ refers to replacing our designed OD Loss with MSE loss. The  $\it w/o\ c$ outperforms other baselines demonstrating the effectiveness of our message passing framework,  while it performs worse than HMOD reflects the need for continuous-time memories. Perhaps due to the fact that ODLoss makes the negative predicted demand acceptable, which makes sense in the real world, it improves the result significantly. Besides, the results on the $\it w/o\ emb$  demonstrate that our OD node embedding plays an important role in predicting OD demand, which could learn an meaningful node embedding.

\section{Conclusion}
This paper aimed to solve the pairwise OD demand prediction problem and proposed a novel dynamic node representation learning framework, HMOD.
A hierarchical memory updater powered by a memory fusion mechanism was proposed to integrate both continuous-time fine-grained spatiotemporal dependencies and discrete-time general spatiotemporal dependencies.
This is the first time traffic node representation learning is extended by continuous-time dynamic graph view.
Meanwhile, an OD node embedding method based on temporal conditioned random walks was also proposed to aggregate semantic information from previous neighbors. 
Experiments on real-world datasets showed high prediction accuracy and robustness of the proposed model.
\subsection*{Acknowledgements}
This work was supported by the National Natural Science Foundation of China (71901011 and U1811463)


\bibliographystyle{named}
\bibliography{ijcai22}

\end{document}